\documentclass[a4paper,sigconf,authorversion,nonacm]{acmart}

\usepackage[utf8]{inputenc}
\usepackage[english]{babel}
\usepackage{tabularx}
\graphicspath{{./images/}}

\AtBeginDocument{%
  \providecommand\BibTeX{{%
    \normalfont B\kern-0.5em{\scshape i\kern-0.25em b}\kern-0.8em\TeX}}}

\acmConference[UrbComp 2021]{The 10th International Workshop on Urban Computing}{November 1, 2021}{Beijing, China}
\acmBooktitle{UrbComp 2021: The 10th International Workshop on Urban Computing, November 1, 2021, Beijing, China}

\begin{document}

\title{Automatic labelling of urban point clouds using data fusion}

\author{Daan Bloembergen and Chris Eijgenstein}
\email{{d.bloembergen, c.eijgenstein}@amsterdam.nl}
\affiliation{%
  \institution{Chief Technology Office, City of Amsterdam}
  \city{Amsterdam}
  \country{The Netherlands}
}

\begin{abstract}
In this paper we describe an approach to semi-automatically create a labelled dataset for semantic segmentation of urban street-level point clouds. We use data fusion techniques using public data sources such as elevation data and large-scale topographical maps to automatically label parts of the point cloud, after which only limited human effort is needed to check the results and make amendments where needed. This drastically limits the time needed to create a labelled dataset that is extensive enough to train deep semantic segmentation models. We apply our method to point clouds of the Amsterdam region, and successfully train a RandLA-Net semantic segmentation model on the labelled dataset. These results demonstrate the potential of smart data fusion and semantic segmentation for the future of smart city planning and management. 
Our code is available on GitHub:
\href{https://github.com/Amsterdam-AI-Team/Urban_PointCloud_Processing}{\color{purple}{\textit{https://github.com/Amsterdam-AI-Team/Urban\_PointCloud\_Processing}}}

\end{abstract}



\maketitle

\section{Introduction}

Semantic segmentation methods offer great potential for automatic understanding of urban street scenes~\cite{ma2018mobile,che2019object} captured by \textit{Mobile Laser Scanning} (MLS) 3D point clouds. Most state-of-the-art methods rely on deep learning algorithms~\cite{zhang2019review} and thus, similar to their 2D counterparts in computer vision, require large amounts of labelled data to train on. However, while computer vision can rely on massive general-purpose training sets that are readily available~\cite{deng2020imagenet, lin2014microsoft, everingham2009the}, such public datasets for 3D semantic segmentation are still mostly lacking.

Some small scale public datasets are available for semantic segmentation, most notably Paris-Lille-3D~\cite{roynard2017parisIJRR} and Toronto 3D~\cite{tan2020toronto3d}, covering a few streets or city blocks. However, there is a lot of diversity in e.g. the building style and the physical appearance of assets such as street lights between different countries or even cities. Moreover, depending on the equipment used, the point density and accuracy might differ significantly between datasets. These factors make that it is not straightforward to use a model trained on one dataset and apply it to a different one and therefore, custom labelled datasets are still needed. Manually annotating point cloud data is a time-consuming task however~\cite{behley2019iccv,tan2020toronto3d}, which for very large-scale point clouds encompassing, e.g., an entire city, quickly becomes infeasible in practice.
Recent works have attempted to alleviate this issue by using topographic maps to automatically label parts of the point cloud~\cite{elberink2020smart,yang2020using}.


In this paper we build on and extend those works. To the best of our knowledge, we present the first \textit{fully modular and open source point cloud processing pipeline}\footnote{\url{https://github.com/Amsterdam-AI-Team/Urban_PointCloud_Processing}} that uses smart data fusion with different data sources to automatically label large parts of urban street-level point clouds. We show that our pipeline can drastically reduce the human time and effort needed to create a fully labelled dataset that is extensive enough to train semantic segmentation models. We highlight this potential by successfully training RandLA-Net~\cite{hu2020randla}, a state-of-the-art deep semantic segmentation model, on the dataset labelled by our method.


\section{Datasets}
 
\subsection{3D point cloud}

Our data of interest is a street-level point cloud of the region of Amsterdam in The Netherlands.\footnote{Point cloud data provided by Cyclomedia: \url{https://www.cyclomedia.com/}} The point cloud is recorded using a panoramic image capturing device together with a Velodyne HDL-32 LiDAR sensor, and thus contains both $(x, y, z)$ coordinates as well as RGB colour information and intensity values.

\begin{figure}[htb]
  \includegraphics[width=\linewidth,trim=0 1.2cm 0 1.5cm,clip]{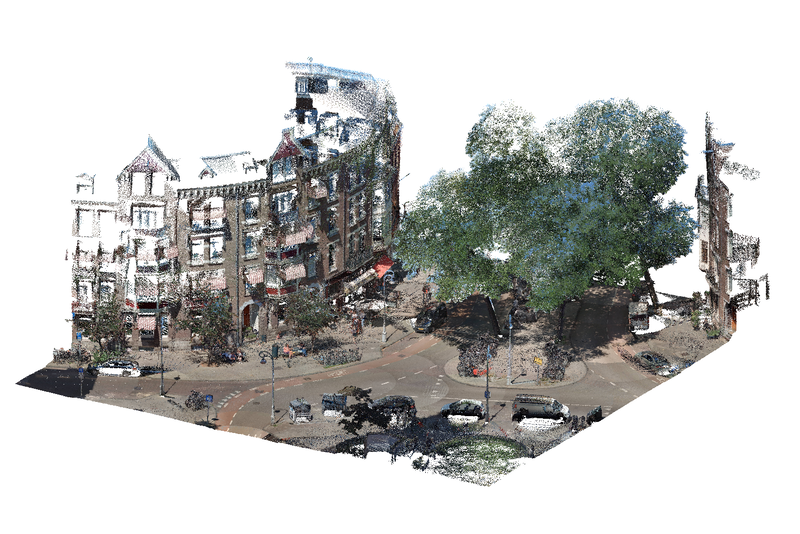}
  \caption{Example point cloud tile.}
  \vspace{-0.5em}
  \label{fig:cloud}
\end{figure}

The point cloud has an average standard deviation of 10cm and a relative precision of 2cm. The point density is dependent on the speed of the mobile laser scanner and the distance to the sensor, and ranges from 1,000 - 2,500 points/m$^2$. The point cloud is split into tiles of 50x50m, which depending on the specific scene can each have up to 15 million points. 
An example tile is shown in Figure~\ref{fig:cloud}, which was visualised using the open source software CloudCompare\footnote{\url{https://www.danielgm.net/cc/}}.

\subsection{AHN elevation data}

Elevation data can provide a rich source of information that can be used for an accurate ground filter, but also to more precisely label buildings, cars, and even pole-like objects as we will see later. We use the AHN3 dataset\footnote{\url{https://www.pdok.nl/introductie/-/article/actueel-hoogtebestand-nederland-ahn3-}} (\textit{Actueel Hoogtebestand Nederland}), the publicly available elevation model of the Netherlands obtained by aerial laser scanning. This data is available as GeoTIFF grid data with a resolution of either 0.5m or 5m, or as a labelled point cloud. We use the latter, as it allows us to extract surface grids with a resolution of 0.1m, as well as height data for buildings. One limitation is that gaps may appear in the elevation data which are caused by, for example, vehicles on the road that obstruct the aerial laser scanner. We include a pre-processing step in which small gaps in the AHN3 data are filled using interpolation.

\begin{figure}[tb]
  \includegraphics[height=5.5cm,trim=0 0.2cm 0 0.8cm,clip]{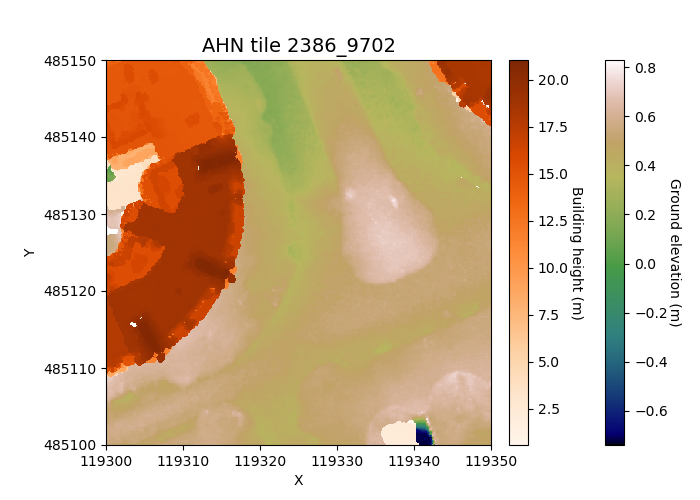}
  \caption{Example AHN3 elevation data showing the ground surface (in green/brown) and building height (in orange/red).}
  \vspace{-0.5em}
  \label{fig:ahn}
\end{figure}

Figure~\ref{fig:ahn} shows an example of the AHN3 elevation data corresponding to the point cloud tile in Figure~\ref{fig:cloud}. One downside of the AHN datasets is their low update frequency; at the moment of writing the AHN3 data for our area of interest is 6 years old.\footnote{\url{https://www.ahn.nl/historie}} This is of particular importance when considering areas that have been recently developed, as buildings might be missing and ground elevation data might have become outdated.

\subsection{BGT topographical map}

\begin{figure}[tb]
  \includegraphics[height=5.5cm]{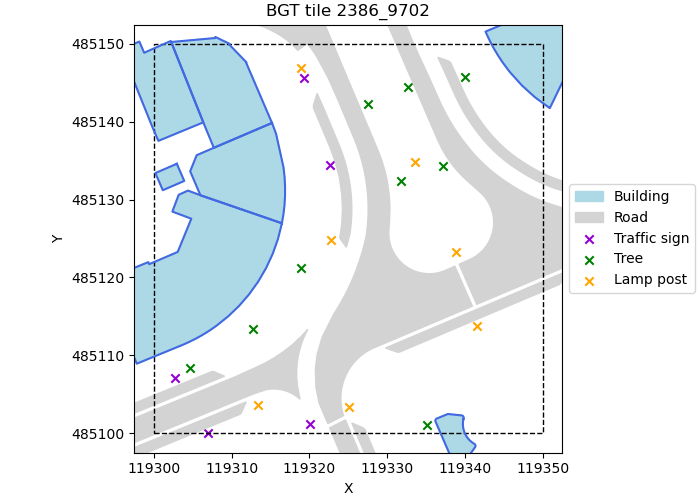}
  \caption{Example BGT data for one point cloud tile.}
  \vspace{-0.5em}
  \label{fig:bgt}
\end{figure}

A second source of rich information is the BGT dataset\footnote{\url{https://www.pdok.nl/introductie/-/article/basisregistratie-grootschalige-topografie-bgt-}} (\textit{Basisregistratie Grootschalige Topografie}), a digital map of large topographical objects which is updated frequently. This map includes building footprint polygons, road part polygons, and $(x, y)$ coordinates of pole-like objects such as trees and lamp posts. We further enrich this using data from the NDW\footnote{\url{https://opendata.ndw.nu/}} (the National Road Traffic Data Portal), in particular we use their dataset of traffic signs (\textit{verkeersborden}). Figure~\ref{fig:bgt} shows the extracted data sources for our example point cloud tile.

\section{Method}

We present a modular pipeline that can be used to label certain objects in the point cloud using data fusion with the various datasets described previously. Each module processes a specific type of information. The modules can be divided into two groups: \textit{data fusion} modules use public data sources to label particular objects in the point cloud, and \textit{region growing} modules further refine the (partially) labelled objects.

Note that the aim of our method is not to correctly label the entire point cloud; instead we only wish to obtain an accurately labelled dataset that can be used to train machine learning models. This means that precision is more important than recall, since we want our training data to be labelled as accurately as possible. Therefore, it is not a big problem if we miss a few objects, as long as the ones that are labelled, are correct.

\subsection{Data fusion}

We use data fusion to label ground, buildings, and cars, and three classes of pole-like objects: trees, lamp posts, and traffic signs.

\paragraph{Ground}

Many urban point clouds consist for a large part of ground, which is why a first pre-processing step in many point cloud classification approaches is to filter out (or label) these ground points in order to simplify further operations~\cite{che2019object}. The typical way of extracting ground points is by identifying horizontal planar surfaces and labelling these as ground~\cite{zermas2017fast}. The disadvantage is that such methods are computationally expensive, and tend to work less well when the ground surface is not flat, or consists of multiple levels.
In order to overcome these limitations we use the AHN3 elevation data as a target surface, and match this with the point cloud. Points in the point cloud which are within a certain margin of the AHN surface (e.g. +/- 25 cm) are labelled as ground.

\paragraph{Buildings}

In order to label buildings we use a combination of BGT footprint polygons and AHN elevation data. The reason to use both data sources simultaneously is that they complement each other: the BGT provides accurate and up-to-date 2D information, while the AHN adds a 3D aspect allowing us to take the building height into account as well.

We mark points that are inside each building polygon (in terms of their $(x,y)$ coordinates) as potential building points. Since both the BGT and the point cloud have a certain margin of error, we inflate the building footprint by 50cm to increase the number of points that are included. It can happen that there are points within the footprint that in fact do not belong to the building. A typical example are trees that partially overlap with the footprint where their branches overhang the building’s roof. To prevent such errors, we use the building height data from AHN, if available, and use this as a cut-off (with a 25cm margin of error) in the $z$ coordinates. Thus, we end up labelling all points that fall within the footprint in terms of $(x,y)$, and are below the building’s roof in terms of $z$. 

\paragraph{Cars}

Cars are relatively easy to detect once the ground has been filtered out, since they have very regular shapes. Using road part polygons from the BGT data (Figure~\ref{fig:bgt}), it is possible to search in a specific location. Using typical car dimensions\footnote{See e.g. \url{https://www.automobiledimension.com/}}, we search for clusters whose minimum bounding rectangle and height match the expected shape and are located above a road part or parking bay.

\paragraph{Pole-like objects}

For the classes trees, lamp posts, and traffic signs we perform a targeted search based on the available BGT data. For each object in the BGT, we extract a small square area surrounding the object's expected $(x, y)$ location (+/-1.5m) from the point cloud. We bin the search area into a 2D grid, and compute statistics for each cell regarding the minimum, maximum, and mean $z$ value within that cell. This allows us to search for pole-like objects.

If we find a match within a maximum distance from the expected location, we additionally compute the radius (thickness) of the object. If the radius falls within an expected range (<0.2m for lamp posts and traffic signs, and <0.5m for trees) we label all points within a cylinder of that radius and at that location as the corresponding object. These points will be used as initial seed points for further refinement of the labelling, as described in the following section.

\subsection{Region growing}

\paragraph{Buildings}

Certain protruding elements such as balconies, bay windows, and canopies are not contained in the building footprint polygons. In order to include these elements, we use a region growing technique. One option is \textit{point-based region growing}~\cite{pcl::RegionGrowing}, in which for each point features are computed in order to determine whether that point should be included or not. This method is very precise, but also computationally expensive.

Since we have a very large number of point cloud tiles to process, we opt for the more efficient approach of \textit{cluster-based region growing}, similar to octree-based approaches~\cite{vo2015octree}. This method does not make a decision for each individual point, but instead it first clusters the point cloud using CloudCompare's octree-based \textit{Label Connected Components} method which is accessible through a Python wrapper.\footnote{\url{https://github.com/tmontaigu/CloudCompare-PythonPlugin}} We then label an entire cluster as building if the fraction of that cluster that was already labelled previously exceeds a threshold, e.g. 0.5. To improve the accuracy of this approach, we use different settings for this threshold, as well as for the octree level in the connected component search, for different parts of the building facade: a more cautious approach near the ground level where there is more clutter; and a coarser approach near the roof, where the point density is lower.

\paragraph{Pole-like objects}

We similarly apply cluster-based region growing to the pole-like objects using the initial seed points that were extracted previously. Again we use different settings for the lower and upper parts of the object; especially the lower part can be cluttered since it is quite common to find bicycles parked against such objects in Dutch cities.

Settings for these region growing modules are elaborate and should be tuned carefully for the dataset at hand. Details about these as well as the data fusion modules can be found in the Jupyter notebooks\footnote{\url{https://github.com/Amsterdam-AI-Team/Urban_PointCloud_Processing/tree/main/notebooks}} accompanying our GitHub repository.

\section{Results}

We now present our results. First, we show that our proposed data fusion pipeline can successfully label point cloud tiles, and we discuss situations in which the labelling fails. Then, we demonstrate the potential of our method by training RandLA-Net, a deep learning semantic segmentation model, on the labelled training set.

\subsection{Automatic labelling example}

\begin{figure}[htb]
  \includegraphics[width=\linewidth,trim=0 1.2cm 0 1.5cm,clip]{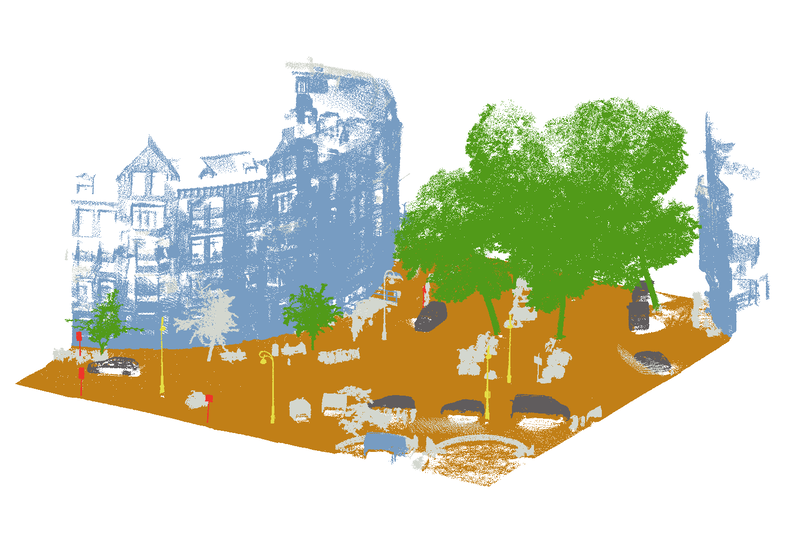}
  \caption{Automatically labelled point cloud tile with ground (brown), buildings (blue), cars (grey), trees (green), lamp posts (yellow), and traffic signs (red).}
  \vspace{-0.5em}
  \label{fig:cloud_labelled}
\end{figure}

Figure~\ref{fig:cloud_labelled} shows the result of running the full pipeline on the example point cloud. Ground, buildings, and cars are labelled correctly. Some of the pole-like objects are missing, which can be attributed to various factors: the small tree left of the centre was not detected because of its slanted angle; the tree branches in the bottom corner belong to a tree outside the tile area; and the location of the lamp post in the centre was off by more than 1.5m in the BGT data. Each of these issues can be likely be corrected by more careful tuning and tweaking of our data fusion modules. However, for the purpose of creating an accurate training set this is not necessary per se: since unlabelled points are ignored during training, these missing objects can potentially be detected and labelled by the semantic segmentation model afterwards.

\subsection{Creating a fully labelled dataset}

In order to further demonstrate the potential of the proposed approach, we use our pipeline to automatically label a dataset of 109 tiles. We then manually check and correct issues, and compare the relative effort of this endeavour compared to annotating from scratch. For manual labelling, we use CloudCompare.\footnote{For a demo, see \url{https://youtu.be/Y27QzOul8WU}}

For our dataset, we selected tiles representing different urban scenes: old town, roads, new development, industrial areas, residential, and high rise. Our manual inspection of the results leads to several observations. 1) Ground and buildings are mostly labelled correctly, with the exception of recently developed areas where the elevation data is outdated. The accurate labelling of ground points also makes it easy to quickly discard reflection noise below ground level. 2) Despite the relatively simple rules, the car labelling module works surprisingly well. Due to our approach, cars that are parked outside official roads and parking bays are not labelled. In addition, some containers placed in parking bays are incorrectly labelled as car. 3) Trees are represented quite accurately in the BGT dataset for our area of interest, and as a result these are also largely labelled correctly. The exception here are trees that are slanted, or trees on private or unmanaged property which are not included in the BGT. 4) Lamp posts and traffic signs are less accurately represented in the BGT dataset. As we opted for precision over recall, objects for which the location error in the BGT is too large are ignored by our approach. In addition, temporary traffic signs around construction areas are missing as these are not included in the BGT. Finally, our approach has difficulty correctly labelling lamp posts or traffic signs that are partially obscured by tree branches or are located very close to buildings or other objects.

Many of these issues can be quickly corrected upon manual inspection, with only a fraction of the effort required to label a full point cloud tile from scratch. In our experience, manually labelling a tile with 10 million points takes roughly 1.5 hours. The main difficulty is labelling the ground, once that is done the remaining points can more easily be clustered. Using our pipeline, we can automatically label the same tile in roughly 20 seconds on a 2020 Intel I7 based laptop. After that, it takes on average 5 minutes to manually inspect the result and correct small mistakes. This substantial reduction in time and effort makes it feasible to create a fully labelled dataset in a matter of hours.

Some statistics of our dataset are presented in Table~\ref{tab:trainset}. Despite including only six classes in our pipeline, just 14\% of the points in our dataset remain unlabelled.

\begin{table}[bt]
  \caption{Class statistics of the labelled dataset.}
  \vspace{-0.5em}
  \label{tab:trainset}
  \begin{tabularx}{\linewidth}{XrXr}
    \multicolumn{4}{l}{\textit{Total number of points: 498,333,348}} \\
    \hline
    Ground & 51.14\% & Tree & 4.65\% \\
    Building & 24.33\% & Lamp post & 0.17\% \\
    Car & 3.74\% & Traffic sign & 0.07\% \\
    Unlabelled & 14.34\% & Noise & 1.53\% \\
    \hline
  \end{tabularx}
\end{table}

\subsection{Semantic segmentation using RandLA-Net}

We split the labelled dataset into training and validation sets of 99 and 10 tiles respectively. We use the training set to train a RandLA-Net semantic segmentation model~\cite{hu2020randla}, and evaluate its performance on the validation set.\footnote{See \url{https://github.com/QingyongHu/RandLA-Net}. We use default values for all parameters, and modify the S3DIS dataloader for our dataset.} As input features we use the $(x, y, z)$ coordinates, and RGB and intensity values normalised to $[0, 1]$. Points that are unlabelled or noise are ignored during training and evaluation. 

Table~\ref{tab:randlanet} shows the performance of the best model found after 100 epochs. 
\begin{table}[tb]
  \caption{RandLA-Net evaluation results: mean IoU (Intersection over Union~\cite{tan2020toronto3d}) and per class IoU on the validation set.}
  \vspace{-0.5em}
  \label{tab:randlanet}
  \begin{tabularx}{\linewidth}{X|XXXXXX}
    mIoU & Ground & Building & Car & Tree & Lamp & Sign \\
    \hline
    88.43 & 97.09 & 96.97 & 92.85 & 97.63 & 82.13 & 63.93 \\
  \end{tabularx}
\end{table}
The model achieves excellent results on detecting ground, buildings, cars, and trees. Performance on lamp posts and traffic signs is notably lagging behind. This is likely due to the relatively low number of example points of these classes in the training set (Table~\ref{tab:trainset}), which is a common problem\footnote{See e.g. the leaderboard at \url{https://npm3d.fr/paris-lille-3d}.} that can potentially be solved by labelling more data.

\begin{figure}[tb]
  \includegraphics[width=\linewidth,trim=0 0.5cm 0 0,clip]{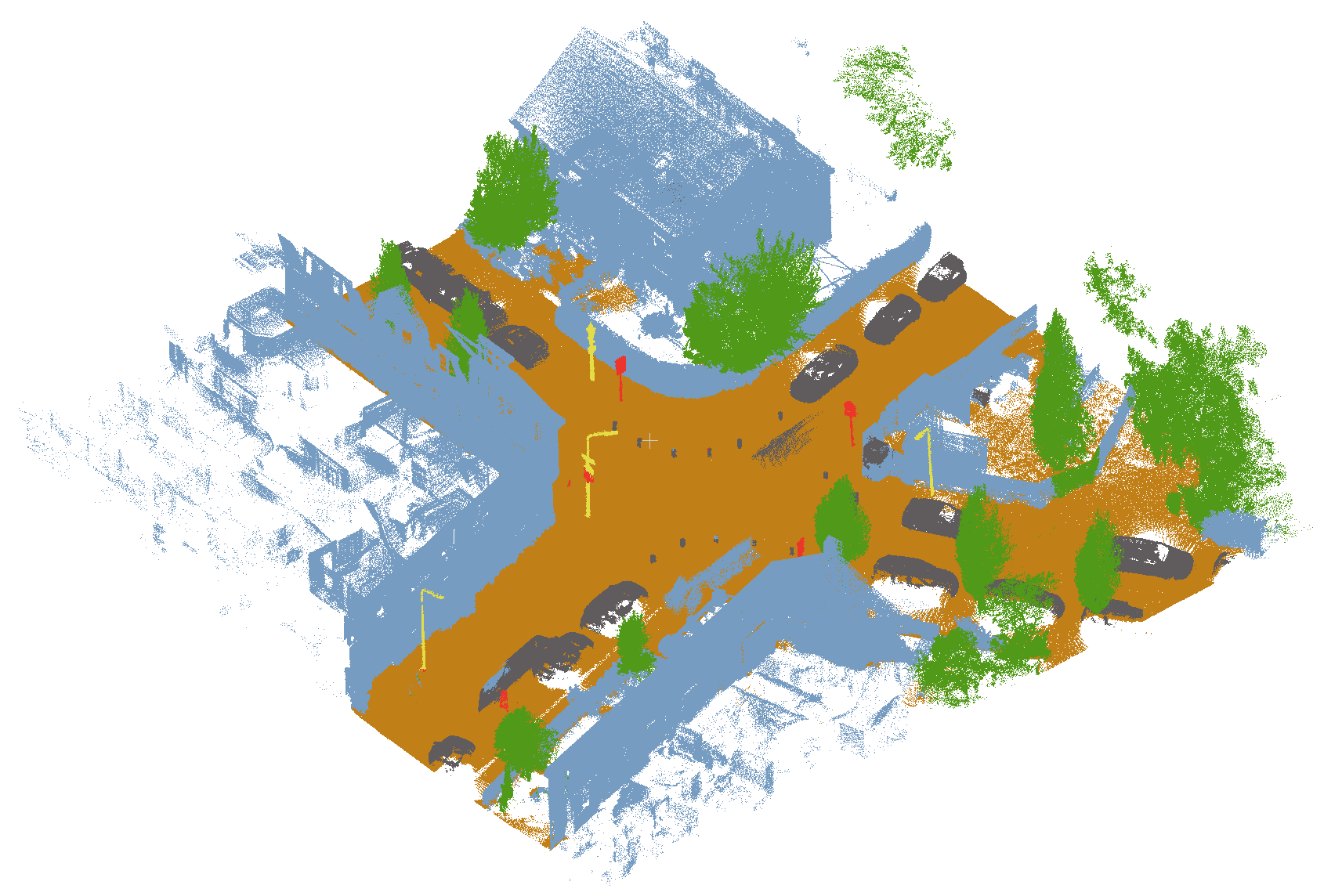}
  \caption{Example tile, not included in the dataset, labelled by the trained RandLA-Net model.}
  \vspace{-1em}
  \label{fig:randlanet_labelled}
\end{figure}

Finally, Figure~\ref{fig:randlanet_labelled} shows an example tile, which was not included in the dataset, that was labelled by the trained RandLA-Net model. This figure confirms the quality of the model; some traffic signs attached to lamp posts are confused, but overall the labels are accurate. Note that objects such as fences and small bollards were not included in the training set, and thus the model cannot distinguish these.

\section{Conclusion \& outlook}

We have demonstrated the potential of using smart data fusion to efficiently label a large dataset that can be used to train deep semantic segmentation models. Our open source modular pipeline is particularly suited for large scale point cloud data with hundreds of millions of points, where full manual annotation is unrealistic and impractical. Evaluation using RandLA-Net highlights the feasibility of our approach, and shows that it is possible to achieve satisfying results while keeping the amount of manual annotation work needed at a manageable level.

Our pipeline can be extended with further modules that label more types of objects of interest, such as bicycle stands, traffic lights, low vegetation, etc. 
In addition, our method can be used for automatic correction of topographical data sources by finding discrepancies between objects' expected locations and their appearance in the point cloud. 
In conclusion, these results demonstrate the potential of semantic segmentation combined with smart data fusion for the automatic understanding and cataloguing of urban street-level point clouds.

\bibliographystyle{ACM-Reference-Format}
\bibliography{refs.bib}

\end{document}